\documentclass[runningheads]{llncs}
\usepackage[T1]{fontenc}
\usepackage{amsmath}
\usepackage{graphicx}
\usepackage{subcaption}
\begin{document}

\title{Chameleon2++: An Efficient and Scalable Variant Of Chameleon Clustering}
\author{Priyanshu Singh\inst{1} \and Kapil Ahuja\inst{1}}
\institute{Math of Data Science and Simulation (MODSS) Lab, Department of Computer Science and Engineering, Indian Institute of Technology Indore, India
\email{priyanshu250167@gmail.com, kahuja@iiti.ac.in}}

\maketitle 

\begin{abstract}
Hierarchical clustering remains a fundamental challenge in data mining, particularly when dealing with large-scale datasets where traditional approaches fail to scale effectively. Recent Chameleon-based algorithms — Chameleon2, M-Chameleon, and INNGS-Chameleon have proposed advanced strategies but they still suffer from $O(n^2)$ computational complexity, especially for large datasets. With Chameleon2 as the base algorithm, we introduce Chameleon2++ that addresses this challenge. 

Our algorithm has three parts. \textit{First}, Graph Generation — we propose an approximate $k$-NN search instead of an exact one, specifically we integrate with the Annoy algorithm. This results in fast approximate nearest neighbor computation, significantly reducing the graph generation time. \textit{Second}, Graph Partitioning — we propose use of a multi-level partitioning algorithm instead of a recursive bisection one. Specifically we adapt the hMETIS algorithm instead of the FM. This is because multi-level algorithms are robust to approximation introduced in the graph generation phase yielding higher-quality partitions, and that too with minimum configuration requirements. \textit{Third}, Merging - we retain the flood fill heuristic that ensures connected balanced components in the partitions as well as efficient partition merging criteria leading to the final clusters.

These enhancements reduce the overall time complexity to $O(n \log n)$, achieving scalability. On real-world benchmark datasets used in prior Chameleon works, Chameleon2++ delivers an average of \textbf{4\%} improvement in clustering quality. This demonstrates that algorithmic efficiency and clustering quality can co-exist in large-scale hierarchical clustering.

\keywords{Hierarchical clustering \and Approximate nearest neighbors \and Multilevel graph partitioning \and Computational complexity \and Scalability}

\end{abstract}

\section{Introduction}
Hierarchical clustering represents a cornerstone technique in data mining and machine learning, providing intuitive tree-structured representations of data relationships that are essential for exploratory data analysis, pattern recognition, and knowledge discovery. As modern applications generate increasingly large-scale and high-dimensional datasets, the computational demands of traditional hierarchical clustering algorithms have become a critical bottleneck, limiting their applicability to real-world scenarios where scalability is paramount.

The Chameleon family of algorithms has emerged as a leading approach to hierarchical clustering. These algorithms consist of three components: graph generation, graph partitioning, and merging. Variations in these components have led to recent advances such as Chameleon2 (Ch2)~\cite{barton2019chameleon2}, M-Chameleon (M-Ch)~\cite{zhang2021chameleon}, and INNGS-Chameleon (INNGS-Ch)~\cite{zhang2021chameleon_inng}. However, these algorithms universally suffer from $O(n^2)$ computational complexity, which we reduce. While this is acceptable for small and synthetic datasets, it becomes a challenge when working with large datasets — a limitation we address here.

We introduce Chameleon2++ (Ch2++), a extension of the Ch2 algorithm specifically designed to address these computational limitations while maintaining clustering quality. Our approach systematically enhances fundamental stages of the Chameleon framework. First, instead of exact $k$-NN computation, we adapt the Annoy (Approximate Nearest Neighbors Oh Yeah)~\cite{aumuller2020ann} algorithm for approximate nearest neighbors (ANN) search, dramatically reducing graph construction time without significant quality degradation. 

Second, existing Ch2 uses recursive Fiduccia-Mattheyses (FM)~\cite{fiduccia1988linear} bisection for graph partitioning, which works well on synthetic datasets and not on real-world scenarios due to its high configurability (requires extensive manual tuning). Instead, we integrate the multilevel graph partitioning (hMETIS)~\cite{karypis1997multilevel} algorithm for graph partitioning, which produces higher-quality partitions with minimal configuration requirements for large-scale and high-dimensional datasets. This also has an added advantage of, superior robustness to approximation-induced noise in the graph structure.

Third, from Ch2 we retain the flood-fill heuristic for achieving connected and balanced components of partitions and as well as the partition merging criteria leading to final clusters. With these changes, Ch2++ achieves $O(n \log n)$ computational complexity compared to the original $O(n^2)$.

Next, we perform experiments on all the real-world benchmark datasets utilized in Ch2, M-Ch and INNGS-Ch. We achieve clustering performance improvements of 1\%, 8\%, and 3\% over these respective algorithms. On an average, we observe a gain of 4\% in clustering performance across all large-scale and high-dimensional benchmark datasets. These results establish that algorithmic efficiency and clustering quality can indeed coexist in large-scale hierarchical clustering applications.

The remainder of this paper is organized as follows: Section \ref{Ch2++:Algorithm} presents our proposed algorithm, Ch2++. Section \ref{Ch2++:Results} reports experimental results and comparisons. Finally, Section \ref{Ch2++:Conclusion} concludes the paper and outlines future directions.

\section{Chameleon2++: Algorithm Design}\label{Ch2++:Algorithm}
This section introduces our scalable and efficient variant of the chameleon algorithm, describing its phases of graph generation, graph partitioning, and merging in the respective subsections below, followed by a summary and complexity analysis. 

\subsection{Graph Generation: Annoy}
In our algorithm, we generate an approximate $k$-NN graph using Annoy, unlike the exact $k$-NN computation used in Ch2. Annoy is one of the simple and effective algorithm in the category of approximate nearest neighbors search and is publicly available \cite{Annoy-spotify}.

\vspace{-0.7cm} 
\begin{figure}[h]
\centering
\includegraphics[width=0.4\columnwidth]{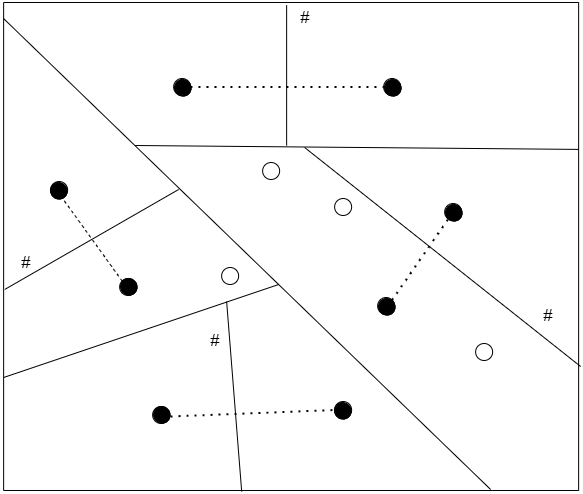}
\hfill
\includegraphics[width=0.575\columnwidth]{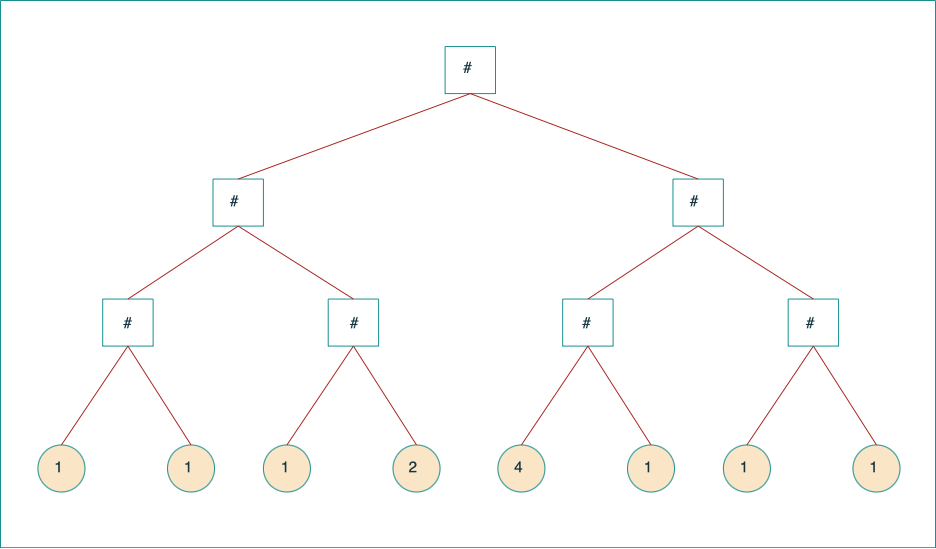}
\caption{The figure on the left illustrates the recursive bi-partitioning of the search space, while the figure on the right shows the tree-based index built for search, in Annoy.}
\label{Annoy}
\end{figure}
\vspace{-0.7cm}

In this algorithm, initially, two random data points are selected, and a binary spaced partitioning is performed based on a selected hyperplane. The hyperplane here is the perpendicular bisector of the line segment connecting the two selected data points. This process is recursively performed, where the data points are chosen from the subset of data being partitioned at each step, as illustrated in Fig. \ref{Annoy}. This continues until each partition has $\le l$ items (here, $l$ is leaf-size parameter).

Next, a Random Projection tree (RP-tree) is built, which serves as the indexing mechanism for nearest neighbor search. The internal nodes here represent the hyperplane and the leaf-nodes denote the search subspace. Let $q$ be the query data point for which the $k$-nearest neighbors are required. Starting with the root of a tree, the path to that child of the root is traversed, which is closer to $q$. This process is repeated until we reach a leaf node. Building of a sample RP-Tree is shown in Fig. \ref{Annoy}. Here, the number inside the node denotes the count of data points present in that subspace.

When we perform recursive bi-partitioning of the search space, the points are chosen at random. To improve the accuracy and search performance of the algorithm this partitioning is performed in-parallel, with different sets of initial random points, and subsequently multiple RP-trees are built i.e., a forest with $(t)$ trees. The search is done simultaneously across all trees using a priority queue-based traversal. Finally, the union of all the data points obtained from the leaf nodes of all the trees are taken into consideration (after removing duplicates). This gives us the final search subspace for the $k$-nearest neighbors for $q$.

The time complexity of Annoy is $O(d \cdot t \cdot n \log n)$, where $d$ is the number of dimension of data, $t$ is the number of trees and $n$ is the number of points in the dataset. Since, $d$ and $t$ are significantly smaller than $n$ ($d, t \ll n$), the overall time complexity is bounded by $O(n \log n)$ \cite{yan2019k}, \cite{zhao2024approximate}, \cite{medium2020knn}.

\subsection{Graph Partitioning: hMETIS}
In our algorithm for graph partitioning, we use hMETIS as compared to FM which is used in Ch2. hMETIS developed in \cite{karypis1997multilevel}, is a multilevel partitioning algorithm which consists of three key phases: coarsening, initial partitioning, and uncoarsening (refinement).

\vspace{-0.6cm}
\begin{figure}[h]
    \centering
    \begin{subfigure}[t]{0.43\columnwidth}
        \includegraphics[width=\linewidth]{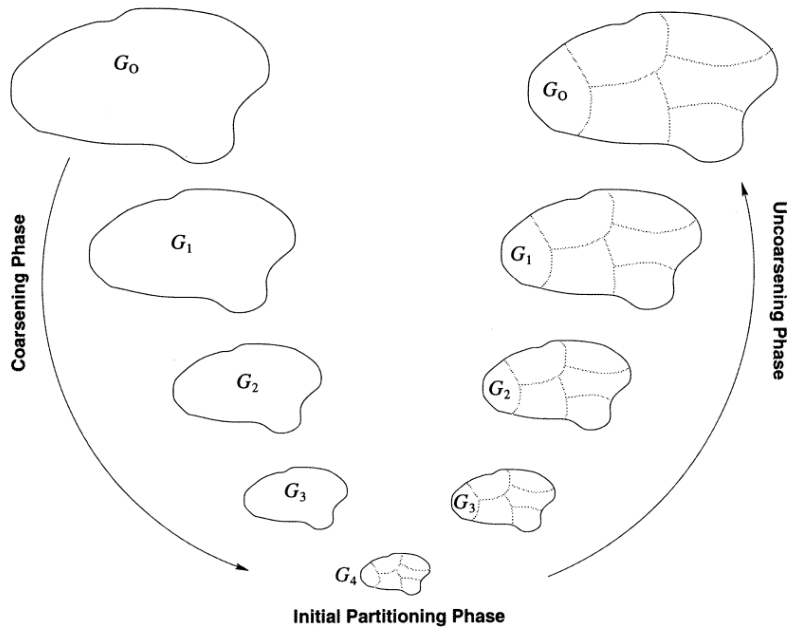}
        \caption{The multilevel paradigm of the hMETIS partitioning algorithm.}
        \label{fig:hmetis}
    \end{subfigure}
    \hfill
    \begin{subfigure}[t]{0.53\columnwidth}
        \includegraphics[width=\linewidth]{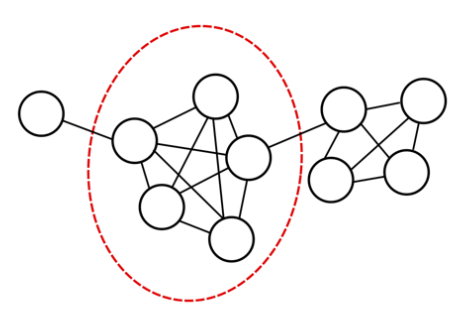}
        \caption{Need for partition refinement: red border shows disconnected vs connected partitions.}
        \label{fig:bfspp}
    \end{subfigure}
    \caption{hMETIS and Flood-Fill.}
    \label{fig:hMETISBFS++}
\end{figure}
\vspace{-0.6cm}

\begin{enumerate}
    \item \textbf{Coarsening Phase} -
    In this phase, hMETIS iteratively creates a sequence of smaller graphs by collapsing edges and fusing nodes. The algorithm uses various matching schemes to identify sets of nodes to be combined. For instance, it combines the nodes which have large number of edges between them. This process continues until the graph is sufficiently small for initial partitioning.

    \item \textbf{Initial Partitioning Phase} -
    Once the graph is coarsened to a manageable size, hMETIS applies a fundamental partitioning algorithm (such as Kernighan–Lin) to create an initial partition. This step is crucial as it provides a starting point for the subsequent refinement process.

    \item \textbf{Uncoarsening Phase} -
    In this phase, hMETIS gradually expands the partitioned graph back to its original size. At each level, the algorithm applies refinement techniques to improve the partition quality. This process is illustrated in Fig. \ref{fig:hmetis}.
\end{enumerate}

As mentioned before, compared to FM, hMETIS is more robust to the approximation errors introduced in the graph generation phase. The reason for this is that hMETIS employs multiple optimization techniques during the partition refinement process, in-turn often avoiding the local optima.

hMETIS has a runtime complexity of $O(n + m \log m)$, where $n$ is the number of vertices and $m$ is the number of partitions \cite{karypis1999hierarchical}.

\subsection{Merging}
A major issue with partitioning algorithms is their tendency to create disconnected partitions. These widely separated partitions lack internal connectivity, which cannot be corrected in later stages, resulting in incorrect merges during the merging phase and ultimately degrading overall clustering quality. Fig. \ref{fig:bfspp} illustrates this issue.

To address this, we employ a local breadth-first search (BFS) based heuristic called Flood-Fill, which has been proposed by Ch2. The algorithm separates them into disjoint connected partitions. Here, the run-time complexity of the heuristic is $O(n)$ \cite{barton2019chameleon2}.

The final phase of the algorithm is merging, where partitions formed after flood-fill are combined into the final clusters. While most existing methods rely solely on external properties such as inter-partition distance, internal properties like density and structural similarity are equally crucial for achieving high-quality clustering. Our merging strategy draws inspiration again from Ch2 \cite{barton2019chameleon2} (which is further inspired by \cite{shatovska2012modified}). This strategy employs two similarity metrics: Relative-Interconnectivity ($R_{IC}$) and Relative-Closeness ($R_{CL}$). $R_{IC}$ quantifies the external interconnectivity between partitions relative to their internal interconnectivity, while $R_{CL}$ captures external closeness relative to internal closeness. This ensures that merging considers both external affinity and internal cohesion, resulting in more coherent final clusters.

Finally, the similarity criteria used for merging partitions is a polynomial combination of $R_{IC}$ and $R_{CL}$. At each iteration, our algorithm selects the partition pairs that maximize this similarity criteria in a agglomerative (bottom-up) manner. Ultimately, the algorithm reports the clustering achieved with highest clustering metric at any merging stage. Since, we are merging $m$ partitions, the time complexity of this is $O(m^2\log{m})$ \cite{bruna2015chameleon}.

\subsection{Summary and Complexity Analysis}
We term our overall algorithm with the above three components as Chameleon2++ (Ch2++). The complexity analysis for each phase is presented in Table~\ref{Ch2++}. With the modifications discussed, the complexity of Ch2++ is \( O(n \log n + n + m\log m + n + m^2 \log m) \).

\begin{table}[h]
\caption{Phases of the Ch2++ algorithm and their computational complexities.}
\centering
\small
\begin{tabular}{|l|l|c|}
\hline
\textbf{Phase} & \textbf{Algorithm} & \textbf{Complexity} \\ \hline
Graph Generation & Approx. k-NN (Annoy) & $O(n\log n)$ \\ \hline
Graph Partitioning & hMETIS & $O(n+m\log m)$ \\ \hline
Merging & Flood-Fill | $R_{CL}^{\alpha} \cdot R_{IC}$ & $O(n)$ + $O(m^2\log m)$ \\ \hline
\end{tabular}
\label{Ch2++}
\end{table}

Given that the number of initial partitions is significantly smaller than the number of data points (i.e., \( m \ll n \); see the results section for an empirical choice of $m$). Hence, the overall complexity is dominated by \( O(n \log n) \), enabling efficient scalability for large-scale clustering and improving over recent Chameleon variants.

\section{Results} \label{Ch2++:Results}
In this section, we present a comprehensive evaluation of Ch2++ against three recent advancements in Chameleon clustering variants: Ch2, M-Ch, and INNGS-Ch. To demonstrate the robustness and scalability of our approach, our evaluation is conducted on real-world datasets, which are large-scale and high-dimensional in nature, and were previously utilized by these recent works. The results are organized into four subsections: Section \ref{Ch2++:Params} gives the parametric configurations of Ch2++, Section \ref{Ch2++:Ch2} presents the performance comparison with Ch2, Section \ref{Ch2++:M-Ch} evaluates Ch2++ against M-Ch, and Section \ref{Ch2++:INNGS-Ch} compares our method with INNGS-Ch.

\subsection{Ch2++ Parametric Configurations}\label{Ch2++:Params}
Here, we present the parameter values for Ch2++. The defaults are listed in Table \ref{tab:ch2pp-params}. Columns 1–3 represent the parameter name, its description, and the default value used in our implementation, respectively.

\vspace{-1.0cm}
\begin{table}[htbp]
\centering
\caption{Default Ch2++ Parameters. Here, $n =$ dataset size.}
\label{tab:ch2pp-params}
\begin{tabular}{|l|l|l|}
\hline
\textbf{Parameter} & \textbf{Description} & \textbf{Value} \\
\hline
$k$       & Approximate k-NN & $r \cdot \log n$ \\
\cline{3-3}
          &                  & $r \in \{1, 2, 4, 8\}$ \\
\cline{3-3}
          &                  & $\log{_\textit{base}} \in \{\ln, \log_{10}, \log_{2}\}$ \\
\hline
$t$       & \#Trees (Annoy)  & $k$ \\
\hline
$l$       & Leaf-Size (Annoy) & $k$ \\
\hline
$m$       & \#Partitions (hMETIS) & $\sqrt{n}/2$ \\
\hline
$m_{\text{fact}}$ & Factor For Small Partitions & $10^3$ \\
\hline
$\alpha$  & $R_{CL_2}$ Priority & 2.0 \\
\hline
$\beta$   & $R_{IC_2}$ Priority & 1.0 \\
\hline
\end{tabular}
\end{table}
\vspace{-0.6cm}

As evident from the table, for experimentation, we use 12 combinations of $k$ (derived from $r \in {1, 2, 4, 8}$ and $\log{\textit{base}} \in {\ln, \log{10}, \log_{2}}$) in $k$-NN graph generation, while keeping the remaining parameters constant, as listed in the table. Finally, we report the best clustering performance achieved across the 12 combinations for each dataset. Notably, here we set $m = \sqrt{n}/2$ in our implementation, hence $(m \ll n)$.

\subsection{Comparison with Chameleon2} \label{Ch2++:Ch2}
For comparison with Ch2, we employ the Normalized Mutual Information (NMI)~\cite{kvalseth2007entropy} metric. NMI measures the agreement between clustering labels and ground truth, normalized to the range [0,1], i.e.,

\begin{equation}
\text{NMI} = \frac{2\,I(X;Y)}{H(X)+H(Y)},
\label{eq:nmi}
\end{equation}

\begin{flushleft}
where $I(X;Y)$ is the mutual information between true labels $X$ and predicted labels $Y$, and $H(X)$, $H(Y)$ are the entropies of $X$ and $Y$ respectively. NMI is symmetric, normalized, and insensitive to cluster-size imbalance, making it particularly suitable for evaluating clustering algorithms dealing with varying cluster distributions.
\end{flushleft}

Although Ch2 has experimented with both small toy problems and real-world datasets, we compare against the latter which is the focus here. Table~\ref{tab:ch2_comparison} gives this comparison of Ch2++ with Ch2. Columns~1–7 denote the dataset name, size ($n$), number of features~($d$), scaling factor $(r)$ $\&$ logarithmic base used in computing $k$-NN $(k = r \cdot \log_{\text{base}})$ for Ch2++, NMI values for Ch2++, and NMI values for Ch2, respectively, with the best value for each dataset highlighted in bold.

\vspace{-0.6cm}
\begin{table}[htbp]
\centering
\caption{Clustering NMI comparison between Ch2++ and Ch2 on real-world datasets.}
\label{tab:ch2_comparison}
\begin{tabular}{|l|l|l|l|l|c|c|}
\hline
Dataset & $n$ (size) & $d$ (features) & $r$ & $\log{_\textit{base}}$ & Ch2++ & Ch2 \\
\hline
pendigits        & 10,992 & 16    & 2  & $\log_{2}$  & \textbf{0.88} & 0.87 \\
cytof.h2         & 31,721 & 32    & 8  & $\log_{10}$ & 0.95          & \textbf{0.98} \\
cytof.h1         & 72,463 & 32    & 1  & $\ln$       & 0.95          & \textbf{0.98} \\
cytof.one        & 81,747 & 13    & 2  & $\log_{2}$  & \textbf{0.88} & \textbf{0.88} \\
mnist            & 70,000 & 784   & 2  & $\ln$       & 0.80          & \textbf{0.82} \\
olivetti-faces   & 400    & 4,096 & 2  & $\log_{10}$ & \textbf{0.86} & 0.77 \\
\hline
\textbf{AVG.}    & --     & --    & -- & --          & \textbf{0.89} & 0.88 \\
\hline
\end{tabular}
\end{table}
\vspace{-0.5cm}

As evident from this table, Ch2++ demonstrates superior performance on 2 datasets, equal performance on 1 dataset, and slightly lower performance on 3 datasets. Overall, Ch2++ achieves an average NMI of 0.89 compared to Ch2's 0.88, representing an improvement of \textbf{1\%} while maintaining competitive performance across all tested datasets. 

\subsection{Comparison with M-Chameleon} \label{Ch2++:M-Ch}

M-Ch uses Accuracy (ACC) as a metric for clustering performance evaluation~\cite{xu2018improved}. Accuracy measures the fraction of correctly assigned labels after optimal label matching using the Hungarian algorithm, i.e.,

\begin{equation}
\text{ACC} = \frac{1}{n}\sum_{i=1}^n \delta\bigl(X_i,\;\mathrm{map}(Y_i)\bigr),
\label{eq:accuracy}
\end{equation}

\begin{flushleft}
where the indicator function $\delta$ is defined as
\end{flushleft}

\begin{equation}
\delta(x,y) = \begin{cases}
1 & \text{if } x = y, \\
0 & \text{otherwise}.
\end{cases} \,
\label{eq:delta_function}
\end{equation}

\begin{flushleft}
Here, $X_i$ is the true label, $Y_i$ is the predicted label, $n$ is the total number of samples, and $\mathrm{map}(\cdot)$ is the optimal permutation of cluster IDs. ACC ranges from 0 to 1, with higher values indicating better clustering performance.
\end{flushleft}

Table~\ref{tab:mch_comparison} presents the clustering accuracy results comparing Ch2++ and M-Ch on real-world datasets from the UCI repository. Here, Columns~1–7 denote the dataset name, size~$(n)$, number of features~$(d)$, scaling factor $(r)$ $\&$ logarithmic base used in computing $k$-NN $(k = r \cdot \log_{\text{base}})$ for Ch2++, ACC values for Ch2++, and ACC values for M-Ch, respectively. The best value for each dataset is highlighted in bold.

\vspace{-0.7cm}
\begin{table}[htbp]
\centering
\caption{Clustering ACC comparison between Ch2++ and M-Ch on real-world UCI datasets.}
\label{tab:mch_comparison}
\begin{tabular}{|l|l|l|l|l|c|c|}
\hline
Dataset & $n$ (size) & $d$ (features) & $r$ & $\log{_\textit{base}}$ & Ch2++ & M-Ch \\
\hline
soybean         & 47    & 35  & 4  & $\log_{10}$ & \textbf{0.96} & 0.89 \\
iris            & 150   & 4   & 4  & $\log_{10}$ & \textbf{0.96} & 0.90 \\
wine            & 178   & 13  & 2  & $\log_{2}$  & 0.72          & \textbf{0.73} \\
seeds           & 210   & 7   & 8  & $\ln$       & \textbf{0.88} & 0.73 \\
sonar           & 208   & 60  & 8  & $\log_{10}$ & \textbf{0.59} & 0.57 \\
ionosphere            & 351   & 34  & 8  & $\log_{2}$  & \textbf{0.77} & 0.69 \\
balance-scale   & 625   & 4   & 2  & $\log_{10}$ & \textbf{0.69} & 0.65 \\
wireless-indoor-localization             & 2,000 & 7   & 8  & $\log_{10}$ & \textbf{0.95} & 0.89 \\
\hline
AVG.   & --    & --  & -- & --          & \textbf{0.82} & 0.76 \\
\hline
\end{tabular}
\end{table}
\vspace{-0.4cm}

As evident from the table, Ch2++ outperforms M-Ch on most of the datasets, specifically outperforming M-Ch on 7 datasets and performing slightly worse on 1 dataset. Ch2++ achieves an average accuracy of 0.82 compared to M-Ch's 0.76, representing a substantial improvement of \textbf{8\%}.

\subsection{Comparison with INNGS-Chameleon} \label{Ch2++:INNGS-Ch}

The clustering quality metric used for this comparison is ACC, as defined in Section~\ref{Ch2++:M-Ch}. Ch2++ is evaluated against INNGS-Ch on real-world datasets from the UCI repository. Table~\ref{tab:inngs_comparison} follows same structure as Table~\ref{tab:mch_comparison} except, instead of M-Ch the results for INNGS-Ch are given the last column. Again, here the best value is highlighted in bold.

% \vspace{-0.5cm}
\begin{table}[htbp]
\caption{Clustering ACC comparison between Ch2++ and INNGS-Ch on real-world UCI datasets.}
\label{tab:inngs_comparison}
\centering
\begin{tabular}{|l|l|l|l|l|c|c|}
\hline
Dataset & $n$ (size) & $d$ (features) & $r$ & $\log{_\textit{base}}$ & Ch2++ & INNGS-Ch \\
\hline
soybean         & 47    & 35   & 4  & $\log_{10}$ & \textbf{0.96} & \textbf{0.96} \\
iris            & 150   & 4    & 4  & $\log_{10}$ & \textbf{0.96} & 0.95          \\
balance-scale   & 625   & 4    & 4  & $\log_{10}$ & \textbf{0.69} & 0.57          \\
wireless-indoor-localization             & 2,000 & 7    & 4  & $\log_{10}$ & \textbf{0.95} & 0.96          \\
ecoli           & 336   & 8    & 8  & $\ln$       & \textbf{0.78} & 0.75          \\
dermatology     & 366   & 34   & 4  & $\ln$       & \textbf{0.58} & 0.52          \\
heart           & 270   & 13   & 8  & $\ln$       & 0.54          & \textbf{0.62} \\
pima            & 768   & 8    & 8  & $\ln$       & \textbf{0.66} & 0.68          \\
yeast           & 1,484 & 8    & 4  & $\ln$       & \textbf{0.49} & 0.39          \\
glass           & 214   & 10   & 1  & $\log_{10}$ & \textbf{0.76} & 0.75          \\
\hline
AVG.            & –     & –    & –  & –           & \textbf{0.74} & 0.72          \\
\hline
\end{tabular}
\end{table}
% \vspace{-0.7cm}

As evident from the results, Ch2++ performs better on 6 datasets, achieves equal performance on 1 dataset, and performs slightly worse on 3 datasets. Overall, Ch2++ achieves an average accuracy of 0.74 compared to INNGS-Ch's 0.72, representing an improvement of \textbf{3\%}.

\section{Conclusion} \label{Ch2++:Conclusion}
We present Ch2++, the first Chameleon-based hierarchical clustering algorithm to break the $O(n^2)$ complexity barrier while maintaining superior clustering quality. Our algorithm achieves $O(n \log n)$ computational complexity through strategic enhancements such as; adapting Annoy's approximate nearest neighbors search instead of exact $k$-NN computation; leveraging multilevel hMETIS for robust graph partitioning instead of use of the FM algorithm; and retaining flood-fill and efficient Ch2 merging criteria.

Experimental evaluation demonstrates consistent performance improvements across all three recent Chameleon variants: 1\% over Ch2, 8\% over M-Ch, and 3\% over INNGS-Ch, on real-world datasets which are large-scale and high-dimensional. Our comprehensive evaluation demonstrates that approximate $k$-NN with hMETIS partitioning does not deteriorate clustering performance, establishing that algorithmic efficiency and clustering quality can coexist.

Future work includes integrating advanced pool of approximate nearest neighbors techniques such as HNSW graphs, LSH, and others \cite{abbasifard2014survey} for enhanced efficiency in sparse high-dimensional datasets, and comprehensive evaluation on ultra-large-scale datasets ($>1$M points). Other directions include integrating compressed sensing methods \cite{agrawal2021csis}, applying optimization techniques to spectral clustering and leveraging the same \cite{shastri2019vector}, developing numerical analysis of algorithmic stability \cite{choudhary2018stability}, and deploying Ch2++ in complex information systems \cite{kim2005effectiveness}.


\begin{thebibliography}{18}

\bibitem{barton2019chameleon2}
Barton, T., Bruna, T., Kordik, P.: Chameleon2: An improved graph-based clustering algorithm. ACM Transactions on Knowledge Discovery from Data 13(1), 1–27 (2019).

\bibitem{zhang2021chameleon}
Zhang, Y., Ding, S., Wang, L., Wang, Y., Ding, L.: Chameleon algorithm based on mutual k‑nearest neighbors. Applied Intelligence 51(4), 2031–2044 (2021).

\bibitem{zhang2021chameleon_inng}
Zhang, Y., Ding, S., Wang, Y., Hou, H.: Chameleon algorithm based on improved natural neighbor graph generating sub-clusters. Applied Intelligence 51(11), 8399–8415 (2021).

\bibitem{aumuller2020ann}
Aumüller, M., Bernhardsson, E., Faithfull, A.: ANN-Benchmarks: A benchmarking tool for approximate nearest neighbor algorithms. Information Systems 87, 101374 (2020).

\bibitem{fiduccia1988linear}
Fiduccia, C.M., Mattheyses, R.M.: A linear-time heuristic for improving network partitions. In: Papers on Twenty-Five Years of Electronic Design Automation, pp. 241–247 (1988).

\bibitem{Annoy-spotify}
W. Li, Y. Zhang, Y. Sun, W. Wang, M. Li, W. Zhang, and X. Lin, ``Approximate nearest neighbor search on high dimensional data—experiments, analyses, and improvement,'' \emph{IEEE Transactions on Knowledge and Data Engineering}, vol. 32, no. 8, pp. 1475--1488, 2019. [Online Available]: https://github.com/spotify/Annoy

\bibitem{yan2019k}
Yan, D., Wang, Y., Wang, J., Wang, H., Li, Z.: K-nearest neighbor search by random projection forests. IEEE Transactions on Big Data 7, 147–157 (2019).

\bibitem{zhao2024approximate}
Zhao, J., Both, J.P., Konstantinidis, K.T.: Approximate nearest neighbor graph provides fast and efficient embedding with applications for large-scale biological data. NAR Genomics and Bioinformatics 6, lqae172 (2024).

\bibitem{medium2020knn}
K-Nearest Neighbors Computational Complexity. https://medium.com/data-science/k-nearest-neighbors-computational-complexity-502d2c440d5.

\bibitem{karypis1997multilevel}
Karypis, G., Aggarwal, R., Kumar, V., Shekhar, S.: Multilevel hypergraph partitioning: Application in VLSI domain. In: Proceedings of the 34th Annual Design Automation Conference, pp. 526–529 (1997).

\bibitem{karypis1999hierarchical}
Karypis, G., Han, E.-H., Kumar, V.: A hierarchical clustering algorithm using dynamic modeling. Retrieved from the University Digital Conservancy: https://hdl.handle.net/11299/215363
 (1999).

\bibitem{shatovska2012modified}
Shatovska, T.B., Onoprienko, O.I., Fedorov, A.O.: A modified multilevel approach to the dynamic hierarchical clustering for complex types of shapes 2(11), 11–14 (2012).

\bibitem{bruna2015chameleon}
Brůna, T.: Implementation of the Chameleon Clustering Algorithm. Bachelor's thesis, Czech Technical University in Prague, Faculty of Information Technology, Department of Software Engineering. https://dspace.cvut.cz/handle/10467/62907
 (2015).

\bibitem{kvalseth2007entropy}
Kvalseth, T.O.: Entropy and correlation: Some comments. IEEE Transactions on Systems, Man, and Cybernetics 17(3), 517–519 (2007).

\bibitem{xu2018improved}
Xu, X., Ding, S., Shi, Z.: An improved density peaks clustering algorithm with fast finding cluster centers. Knowledge-Based Systems 158, 65–74 (2018).

\bibitem{abbasifard2014survey}
Abbasifard, M.R., Ghahremani, B., Naderi, H.: A survey on nearest neighbor search methods. International Journal of Computer Applications 95(25) (2014).

\bibitem{agrawal2021csis}
Agrawal, R., Ahuja, K.: CSIS: Compressed sensing-based enhanced-embedding capacity image steganography scheme. IET Image Processing 15(9) (2021).

\bibitem{shastri2019vector}
Shastri, A.A., Ahuja, K., Ratnaparkhe, M.B., Shah, A., Gagrani, A., Lal, A.: Vector quantized spectral clustering applied to whole genome sequences of plants. Evolutionary Bioinformatics 15, 1176934319836997 (2019).

\bibitem{choudhary2018stability}
Choudhary, R., Ahuja, K.: Stability analysis of bilinear iterative rational Krylov algorithm. Linear Algebra and its Applications 538, 56–88 (2018).

\bibitem{kim2005effectiveness}
Kim, S., Murthy, U., Ahuja, K., Vasile, S., Fox, E.A.: Effectiveness of implicit rating data on characterizing users in complex information systems. In: Research and Advanced Technology for Digital Libraries, LNCS, vol. 3652, pp. 1–13 (2005).

\end{thebibliography}
\end{document}